\documentclass[letterpaper]{article}
\usepackage[preprint]{aaai2027}
\usepackage[hyphens]{url}
\usepackage{graphicx}
\usepackage{amsmath}
\usepackage{amssymb}
\usepackage{natbib}
\usepackage{caption}
\usepackage{algorithm}
\usepackage{algorithmic}

\usepackage{newfloat}
\usepackage{listings}
\DeclareCaptionStyle{ruled}{labelfont=normalfont,labelsep=colon,strut=off}
\floatstyle{ruled}
\newfloat{listing}{tb}{lst}{}
\floatname{listing}{Listing}

\usepackage{booktabs}

\title{Fovea-JEPA: Overcoming Geometric Blind Spots in V-JEPA for Camera Motion Recognition}
\author{
    Xiangkun Shi,
    Hanhui Li,
    Shiyue Zhang,
    Xiaodan Liang\corresponding
}
\affiliations{
    Sun Yat-sen University
}

\begin{document}

\maketitle

\begin{abstract}
Camera motion recognition requires distinguishing viewpoint changes induced by camera translation, rotation, or zoom from visual variations caused by object motion and scene dynamics. Current approaches often rely on large vision-language or video-language models, but their overreliance on high-level semantic correspondences makes them unreliable in understanding spatio-temporal relationships. Joint-Embedding Predictive Architectures (JEPAs), which learn by predicting changes in spatio-temporal representations (e.g., V-JEPA), offer a promising way to address this issue. Nevertheless, our experiments reveal an inherent limitation: although JEPAs provide a strong baseline, their emphasis on spatio-temporal modeling may underrepresent the geometric cues of some critical regions, causing even broadly applicable augmentation strategies such as temporal reversal to degrade their performance on several motion families. In contrast, weaker standalone geometry-oriented predictors (e.g., VGGT) with the same augmentations improve across all motion families. We term these geometrically underrepresented regions in JEPAs \textbf{\emph{geometric blind spots}} and propose a Fovea-JEPA framework to identify and address them. Specifically, given V-JEPA as the base model and VGGT as a geometry expert, (i) we first analyze their prediction errors to construct pseudo-labels for weakly supervising a blind-spot predictor. (ii) The blind-spot predictor then generates token-wise intervention maps, which localize geometrically underrepresented regions. (iii) Finally, guided by these intervention maps, we adaptively fuse V-JEPA and VGGT features to perform camera-motion recognition. In this way, Fovea-JEPA addresses the geometric blind spot issue, and experimental results on three public datasets validate its effectiveness: Fovea-JEPA improves the state-of-the-art performance by $9.67\%$ on CameraMotionDataset, $8.67\%$ on CameraBench, and $24.38\%$ on ShotBench, respectively. Code and models will be released upon acceptance.
\end{abstract}

\section{Introduction}
\label{sec:introduction}

\begin{figure*}[!t]
    \centering
    \includegraphics[width=\textwidth, height=8cm]
    {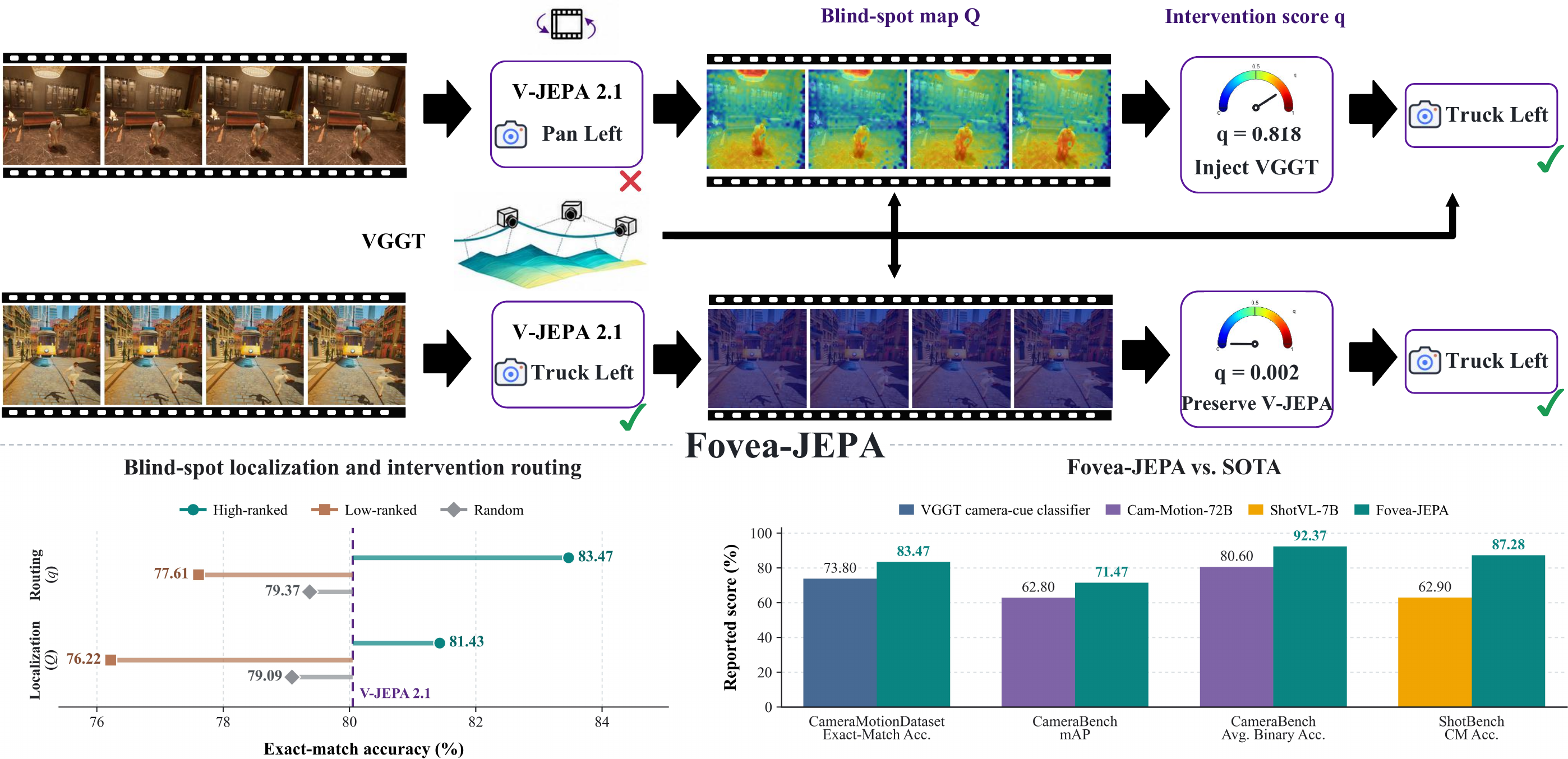}
    \caption{
Motivation of Fovea-JEPA. Top: Fovea-JEPA estimates a blind-spot map $\mathbf{Q}$ and aggregates it into an intervention score $q$, which indicates whether intervention by a geometry expert (e.g., VGGT) is necessary. A high-\(q\) case activates geometric intervention to correct the V-JEPA prediction, whereas a low-\(q\) case preserves the original prediction.
Bottom left: matched-budget intervention controls. Applying intervention to high-\(q\) clips outperforms applying it to low-\(q\) or randomly selected clips, while retaining VGGT
tokens with the highest \(q\) scores outperforms retaining those with the lowest scores or using random masks. Bottom right: Fovea-JEPA compared with state-of-the-art methods across three benchmarks.
}
    \label{fig:teaser}
\end{figure*}

Camera motion recognition is challenging, as it requires identifying how a camera translates, rotates, or zooms while separating viewpoint changes from object motion and scene dynamics. Recent approaches increasingly rely on large vision-language models (VLMs) and VideoLLMs, benefiting from their broad visual-semantic knowledge. However, their reliance on high-level semantic correspondences makes them unreliable in capturing the fine-grained spatio-temporal and geometric relationships required for camera-motion recognition \citep{lin2025camera,feng2026geometry}. Another issue is that geometrically different camera operations can produce similar visual changes. For example, zooming and moving forward can appear similar, while independently moving objects can resemble camera-induced motion. This task therefore calls for representations that model temporal dynamics while preserving discriminative geometric evidence beyond high-level appearance and semantic cues.

Joint-embedding predictive architectures (JEPAs) learn latent representations via spatio-temporal constraints, thereby encouraging robust context modeling \citep{bardes2024vjepa}. For example, compared with contrastive and reconstruction-based video models, V-JEPA \citep{murlabadia2026vjepa21} generates dense features that are robust to visual corruptions and better preserve fine-grained physical distinctions \citep{alrasheed2026latent}. These properties make V-JEPA a strong candidate for camera motion recognition.

However, it remains unclear whether the representational advantages of JEPAs extend consistently across different camera-motion families, particularly when fine-grained geometric evidence is required. To investigate this, we use temporal reversal \citep{pickup2014arrow} as a controlled probe because it reverses direction-dependent motion cues while largely preserving scene appearance and spatial content, thereby revealing whether a representation captures the temporal and geometric evidence needed to distinguish opposing camera motions. We compare the responses of V-JEPA and a VGGT-based predictor specialized in geometric modeling \citep{wang2025vggt}. As reported in Section~\ref{sec:reversal_ablation}, V-JEPA remains the stronger standalone predictor and improves overall under temporal-reversal supervision, yet its accuracy decreases for several motion types. In contrast, the weaker VGGT-based predictor improves across every evaluated motion class. This contrasting phenomenon suggests that VGGT retains complementary geometric cues that V-JEPA underrepresents in some critical regions.

We refer to the above geometrically underrepresented regions in V-JEPA as \textbf{\emph{geometric blind spots}}. Nevertheless, indiscriminately combining V-JEPA and VGGT to handle these blind spots is prone to interfering with reliable V-JEPA predictions. This leads to a critical design requirement of our method, i.e., geometric evidence should complement the primary V-JEPA representation only in regions where are beneficial. We therefore propose Fovea-JEPA, which retains V-JEPA as the primary representation and uses VGGT as a geometry expert to compensate for these blind spots. As illustrated in Figure~\ref{fig:teaser}, Fovea-JEPA learns a token-wise spatio-temporal blind-spot map $\mathbf{Q}$ that localizes where geometric evidence is important, and yields a clip-level intervention score $q$ that determines how the geometry expert should contribute to the final prediction. Through a controlled matched-budget experiment, we validate that high-$q$ regions require and benefit more from geometric intervention. Consequently, Fovea-JEPA selectively applies geometric intervention when it is beneficial and conquers the blind-spot issue. Experimental results on three public datasets show that Fovea-JEPA increases the state-of-the-art performance significantly.

In summary, the contributions of this paper are threefold:
\begin{itemize}
    \item We are the first to apply V-JEPA to camera-motion recognition
    and uncover its geometric blind spot issue.

    \item We propose a Fovea-JEPA framework that adaptively rectifies the visual representation and predictions of V-JEPA via VGGT-based geometric intervention.

    \item We design a weakly supervised blind-spot predictor to generate token-level geometric response maps, which precisely localize visually neglected geometrically salient regions and enable adaptive spatial intervention.

\end{itemize}

\section{Related Work}

\subsection{Camera Motion Understanding}

Current approaches to camera-motion understanding
are predominantly built on VLMs and VideoLLMs. CameraBench
evaluates motion primitives in diverse Internet videos and reveals complementary
limitations of structure-from-motion and VLMs
\citep{lin2025camera}; ShotBench evaluates expert-level shot
language \citep{liu2025shotbench}; and CineTechBench extends
evaluation to broader cinematographic understanding and
generation \citep{wang2025cinetech}. CamReasoner trains
VideoLLMs to perform structured ego-motion inference
\citep{wu2026camreasoner}. Taking a different route from
semantic modeling, Geometry-Guided Camera Motion Understanding
injects 3D-derived motion cues into VideoLLMs to strengthen
their sensitivity to camera geometry beyond appearance and
semantic context
\citep{feng2026geometry}.

This geometry-guided direction builds on advances in 3D reconstruction. Classical
structure-from-motion recovers camera poses and scene structure
through optimization \citep{schonberger2016sfm}, whereas learned
models directly predict 3D structure: DUSt3R regresses point maps
\citep{wang2024dust3r}, VGGSfM scales learned multi-view
reconstruction \citep{wang2024vggsfm}, and MonST3R extends such
reasoning to dynamic scenes \citep{zhang2025monst3r}. VGGT
unifies camera, depth, point-map, and tracking prediction in a
feed-forward model \citep{wang2025vggt}, making it a suitable geometry expert for camera-motion
recognition.

\subsection{Video Representation Learning}

Self-supervised video representation learning commonly uses
contrastive objectives or masked prediction.
Contrastive methods learn invariance across clips from the
same video \citep{qian2021cvrl}, while masked video models
reconstruct pixels or predict teacher features
\citep{feichtenhofer2022maest,tong2022videomae,
wang2023mvd}.
Temporal order provides an additional signal for motion
understanding.
Prior work uses forward--backward classification, structured
label reversal, or time-equivariant learning to learn temporal structure and motion-sensitive
representations
\citep{pickup2014arrow,price2019retro,
jenni2021timeequivariant,wei2018arrow,
jenni2020temporal}.
Recent studies also find that VideoLLMs remain weakly
sensitive to temporal reversal
\citep{xue2025arrow,han2026tracing}.
In our work, temporal reversal serves as a shared supervision
signal for comparing how V-JEPA and VGGT respond across
different camera-motion families.

\subsection{Joint-Embedding Predictive Architectures}

I-JEPA introduced joint-embedding prediction of masked target
representations from visible context without pixel reconstruction
\citep{assran2023ijepa}. V-JEPA extends this principle to video through
latent spatio-temporal prediction \citep{bardes2024vjepa}.
V-JEPA~2 scales it to large-scale video pretraining
\citep{assran2025vjepa2}, while V-JEPA~2.1 improves the spatial
organization and temporal consistency of dense video features
\citep{murlabadia2026vjepa21}. A recent systematic study further
reports that V-JEPA variants outperform the evaluated alternative
video models in corruption robustness, fine-grained physical
discrimination, and temporal-direction sensitivity
\citep{alrasheed2026latent}. These findings establish V-JEPA as a strong video
representation, but its potential limitations in representing
fine-grained geometric evidence remain underexplored. We
study this issue within the scheme of camera motion recognition.

\section{Methodology}

\begin{figure*}[t]
    \centering
    \includegraphics[width=\textwidth]
{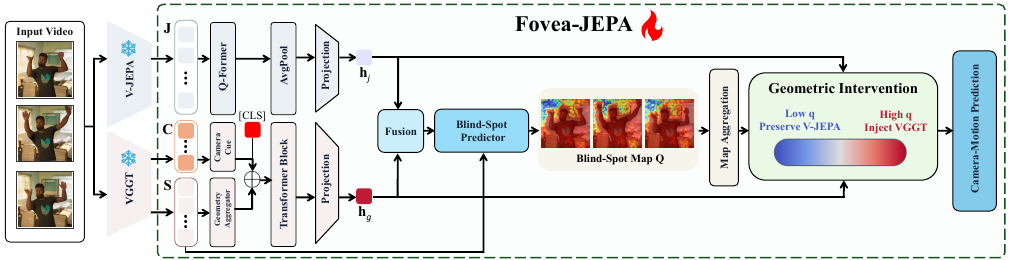}
    \caption{Overview of Fovea-JEPA. Our dual-branch framework leverages visual-geometric complementary cues and hierarchical error-guided pseudo supervision to detect geometric blind spots. Subsequently, our framework adaptively retains robust visual semantic predictions or activates geometric compensation to correct visual prediction errors.}
    \label{fig:framework}
\end{figure*}

\subsection{Overview}
\label{sec:method_overview}
\textbf{Task Formulation.} Let $\mathcal{V}=\{I_t\}_{t=1}^{T}$ denote an input video
with $T$ frames, and let $K$ denote the number of camera-motion
classes. Our goal is to learn a mapping
$F:\mathcal{V}\rightarrow\mathbb{R}^{K}$ that produces
$K$ classification logits.  We implement $F$ with the proposed Fovea-JEPA framework, whose core component is an adaptive geometric intervention paradigm, as illustrated in Figure~\ref{fig:framework}.

Specifically, Fovea-JEPA adopts a dual-branch architecture composed of a JEPA-based visual branch and a geometric branch. Both branches take tokens extracted from the frozen backbones of V-JEPA \citep{murlabadia2026vjepa21} and VGGT \citep{wang2025vggt} as inputs, and produce two feature tokens encoding the visual and geometric context of the video, respectively. The two resulting tokens are fused to form a global context token. Together with dense spatial tokens from VGGT, this global token is fed into a blind-spot predictor to yield a blind-spot map, which highlights regions that are geometrically informative yet neglected by V-JEPA. We subsequently aggregate this blind-spot map into a scalar score to modulate the geometric intervention from VGGT on the global context. Finally, the modulated token is leveraged to predict the target class label.

Despite the intuitive design of the above Fovea-JEPA framework, its practical realization faces substantial challenges, primarily due to the absence of ground-truth annotations for geometric blind spots. A naive fusion of V-JEPA and VGGT features via learnable weights lacks interpretability and may degrade overall model performance, as VGGT exhibits weaker visual reasoning capability compared with V-JEPA. To address these issues, we introduce an error-guided pseudo-label generation strategy for blind-spot estimation in the subsequent section.

\subsection{Error-guided Intervention Labels}

To determine which samples require geometric intervention and produce the corresponding supervision labels, we first assess the recognition performance of the visual branch and the geometric branch. For this purpose, we attach two classification heads as follows:

\textbf{Visual Branch.}
Let $\mathbf{J} \in \mathbb{R}^{D\times T \times N}$ denote the $D$-dimensional visual tokens extracted by the V-JEPA backbone, where $N$ denotes the number of per-frame tokens. Our visual branch aggregates these tokens into a compact visual context token $\mathbf{r}_j \in \mathbb{R}^{D}$ via attention pooling. Specifically, the branch first leverages a Q-Former \citep{li2023blip2} to learn a set of $M$ query vectors $\mathbf{K} \in \mathbb{R}^{D \times M}$. It then performs standard cross-attention between the learned queries and the visual token sequence $\mathbf{J}$. The final visual context token is obtained by averaging the attended features across both spatial and temporal dimensions. This can be formulated as:
\begin{equation}
    \mathbf{A} = \mathrm{Softmax}\left( \mathbf{K}^\top \mathbf{J} \right), \mathbf{r}_j = \mathrm{AvgPool}(\mathbf{J} \mathbf{A}^\top).
\end{equation}
Then, visual classification logits $\mathbf{z}_j \in \mathbb{R}^{K}$ are predicted from $\mathbf{r}_j$ via a lightweight prediction head, formulated as $\mathbf{z}_j = R_j(\mathbf{r}_j)$. Here, $R_j$ denotes the classification head implemented by a multilayer perceptron (MLP).

\textbf{Geometric Branch.} The VGGT backbone produces two types of geometric token sequences for the input video. The first is dense geometric tokens $\mathbf{S} \in \mathbb{R}^{D\times T \times N}$ that preserve full spatio-temporal structural information. The second is compact camera-aware tokens $\mathbf{C} \in \mathbb{R}^{D\times T}$ that encode implicit camera motion priors. For the conciseness of description, we omit the intermediate dimension alignment operations for $\mathbf{S}$ and $\mathbf{C}$, and assume that both geometric tokens share the same feature dimension $D$ and spatial token number $N$ with the visual tokens $\mathbf{J}$.

To enhance the capability of camera tokens in capturing camera motion patterns, we exploit the camera extrinsic matrices estimated by VGGT to extract auxiliary trajectory features. Specifically, we calculate four groups of motion attributes, including the displacement distance between the first and last frame, the magnitude of the translation vector, the frame-wise displacement between adjacent frames, and the corresponding rotation angles. We concatenate these attributes with the extrinsic matrices, forming a 16-dimensional per-frame trajectory feature in total. We then concatenate trajectory features with the original camera tokens $\mathbf{C}$ and adopt an MLP to yield the enhanced camera motion tokens $\mathbf{C}^\ast \in \mathbb{R}^{D\times T}$.

As for the dense geometric tokens $\mathbf{S}$, we adopt an aggregation pipeline similar to that of the visual branch. We use another Q-Former and spatial attention pooling to generate the frame-level geometric feature sequence $\mathbf{r}_s \in \mathbb{R}^{D\times T}$. $\mathbf{r}_s$ is then projected via a lightweight MLP to align its feature space with the enhanced camera tokens $\mathbf{C}^\ast$. The projected geometric representations are subsequently fused with $\mathbf{C}^\ast$ through element-wise addition. The fused token sequence jointly encodes spatial geometric structures and temporal camera motion patterns. To aggregate these frame-level geometric features into a single global geometric context token, we follow the standard ViT paradigm that prepends a [CLS] token to the fused token sequence and feeds the entire sequence into a standard transformer block. The output [CLS] token serves as the final global geometric context $\mathbf{r}_g \in \mathbb{R}^{D}$.

\textbf{Fused Prediction.} To simulate the prediction performance after geometric intervention, we fuse visual and geometric context tokens for joint classification. We first project the visual context $\mathbf{r}_j$ and geometric context $\mathbf{r}_g$ into a shared feature space via two-layer MLPs $P_j$ and $P_g$, yielding aligned features $\mathbf{h}_j$, $\mathbf{h}_g \in \mathbb{R}^{D}$, respectively. $\mathbf{h}_j$, $\mathbf{h}_g$ are aggregated by element-wise addition and then fed into an MLP-based predictor $R_f$ to produce the intervened classification logits $\mathbf{z}_f \in \mathbb{R}^{K}$. This fusion process is formulated as:
\begin{equation}
\mathbf{h}_j = P_j(\mathbf{r}_j),\; \mathbf{h}_g = P_g(\mathbf{r}_g),\; \mathbf{z}_f = R_f(\mathbf{h}_j + \mathbf{h}_g).
\end{equation}

\textbf{Intervention Label Generation.}
To adaptively guide geometric intervention, we construct binary video-level intervention labels based on the prediction consistency between the visual branch and the fused branch. Geometric intervention is necessary for samples where the pure visual branch produces incorrect predictions, while the visual-geometric fusion yields correct results. Let $\widehat{\mathbf{y}}_j^i$ and $\widehat{\mathbf{y}}_f^i$ denote the predicted labels obtained from logits $\mathbf{z}_j$ and $\mathbf{z}_f$ for the $i$-th video sample, and $\mathbf{y}^i$ denote the corresponding ground-truth label. The binary intervention label $y_i^{\mathrm{int}}$ is formally defined as::
\begin{equation}
y_i^{\mathrm{int}}
=
\begin{cases}
1, & \widehat{\mathbf{y}}_j^i\neq\mathbf{y}^i \land \widehat{\mathbf{y}}_f^i=\mathbf{y}^i,\\
0, & \widehat{\mathbf{y}}_j^i=\mathbf{y}^i.
\end{cases}
\label{eq:intervention_label}
\end{equation}
We leave out samples with wrong predictions from both branches without assigning intervention supervision. In this way, $y_i^{\mathrm{int}}=1$ corresponds to visual branch failure cases that can be corrected by geometric complementary cues, requiring adaptive geometric intervention. In contrast, $y_i^{\mathrm{int}}=0$ represents reliable visual predictions that retain accurate semantic information and do not need geometric modulation.

\textbf{Geometrically Salient Regions.}
Although the video-level intervention labels successfully screen out samples requiring geometric compensation, they lack fine-grained spatial awareness and cannot pinpoint regions that dominate performance improvement. To remedy this, we construct token-level supervision targets to identify geometrically salient regions from the geometric branch. We first adopt the frame-averaged geometric token as a neutral baseline for each frame, and leverage first-order gradient cues to quantify the saliency and task contribution of each geometric token:
\begin{equation}
\bar{\mathbf{s}}_{i,t} = \frac{1}{N}\sum_{n=1}^{N}\mathbf{s}_{i,t,n},\; u_{i,t,n} = -\nabla_{\mathbf{s}_{i,t,n}}\mathcal{E}_i \cdot \left( \mathbf{s}_{i,t,n} - \bar{\mathbf{s}}_{i,t} \right).
\label{eq:spatial_reference}
\end{equation}
Here, $\mathcal{E}_i$ denotes the classification loss derived from the geometric branch. A high positive saliency value $u_{i,t,n}$ indicates that the corresponding token resides in geometrically salient regions that carry vital, camera-related information, whose absence will degrade overall prediction performance.

With the above video-level intervention labels and token-level geometric saliency signals, we form a hierarchical complementary intervention label generation pipeline: the former conducts global sample screening to determine the necessity of geometric intervention, while the latter achieves fine-grained localization of geometrically salient regions to support adaptive spatial feature modulation.

\subsection{Blind-Spot Map Estimation}
\label{sec:blind_spot_learning}
Given the video-level intervention labels and fine-grained token-level geometric saliency signals, we construct a blind-spot predictor to jointly infer where spatial geometric intervention is needed and whether global modulation should be activated for each video clip.

We first construct a cross-branch context token that integrates both feature-level and prediction-level statistics to bridge the visual and geometric branches and capture their complementary discrepancies. This can be formulated as,
\begin{equation}
    \mathbf{c}
    =
    \psi_c(
        \mathbf{h}_j,
        \mathbf{h}_g,
        \mathbf{z}_j,
        \mathbf{z}_f
    ).
    \label{eq:cross_branch_context}
\end{equation}
We then employ two linear projection layers, $\phi_s$ and $\phi_c$, to align geometric tokens $\mathbf{S}$ and $\mathbf{c}$ into a shared latent space. The projected $\mathbf{c}$ is broadcast to all spatio-temporal positions of $\mathbf{S}$, and they are fused via element-wise addition. An MLP $F_{\mathrm{bs}}$ then takes the fused features to produce blind-spot logits, which are normalized via a sigmoid function $\sigma$ to generate the final response map $\mathbf{Q}\in[0,1]^{T\times N}$:
\begin{equation}
    \mathbf{Q} =\sigma(F_{\mathrm{bs}}\left(
        \phi_s(\mathbf{S})
        +
        \phi_c(\mathbf{c})
    \right)).
    \label{eq:blind_spot_prediction}
\end{equation}
The blind-spot predictor is learned with the video-level labels and the geometrically salient maps, of which the optimization objectives are provided in Sec. \ref{sec:objectives}.

\subsection{Blind-Spot-Guided Geometric Intervention}
\label{sec:geometry_injection}
For video clips with reliable visual semantics, forced geometric fusion inevitably introduces irrelevant geometric noise and offsets the discriminative visual priors. To address this issue, we propose a blind-spot-guided adaptive geometric intervention strategy, which leverages the predicted blind-spot response map for dual-level adaptive modulation. Namely, the predicted blind-spot map $\mathbf{Q}$ serves two core functions in our framework. First, $\mathbf{Q}$ guides fine-grained token-level feature fusion to activate geometric compensation on salient regions. Second, $\mathbf{Q}$ is aggregated into a global clip-level intervention scalar to determine whether global geometric modulation is required.

\textbf{Clip-Level Intervention Score Aggregation.} To derive a global intervention trigger from the dense blind-spot map $\mathbf{Q}$, we aggregate it into a clip-level scalar intervention score $q$ via a two-stage TopMean pooling strategy. Concretely, we first compute the average response of the top $\xi_1$ salient spatial tokens per frame, and then aggregate the top $\xi_2$ frame-level scores to obtain the final clip-level score $q$. This design inherently filters out noisy and trivial geometric responses, retaining only highly discriminative geometric blind-spot cues.

\textbf{Token-Level Geometric Intervention.} The clip-level score $q$ controls the overall intervention strength, while a context-dependent gate determines the contribution of geometry to each feature channel as follows:
\begin{equation}
\begin{aligned}
    \mathbf{m}
    &=
    q\,\sigma\left(F_{\alpha}(\mathbf{c})\right),\\
    \widetilde{\mathbf{h}}
    &=
    (\mathbf{1}-\mathbf{m})\odot\mathbf{h}_j
    +
    \mathbf{m}\odot\mathbf{h}_g,
\end{aligned}
\label{eq:geometric_intervention}
\end{equation}
and $ \widetilde{\mathbf{z}}=R_f(\widetilde{\mathbf{h}})$. Here, $F_{\alpha}$ is a lightweight modulation network and $\odot$ denotes element-wise multiplication.

At inference, we use an empirical threshold $\tau$ to determine whether the global intervened prediction is necessary:
\begin{equation}
    \mathbf{z}_{\mathrm{out}}
    =
    \begin{cases}
        \widetilde{\mathbf{z}}, & q\geq\tau,\\
        \mathbf{z}_j, & q<\tau.
    \end{cases}
    \label{eq:inference_decision}
\end{equation}
The visual-based prediction is therefore preserved when the
estimated benefit of geometric intervention is low. Otherwise, the intervened prediction is chosen.

\subsection{Optimization Objectives}
\label{sec:objectives}
In this section, we present the learning objectives of Fovea-JEPA. Due to the page limitation, detailed loss definitions are provided in the supplemental material.

Fovea-JEPA is trained with both spatio-temporal and clip-level supervision. For the spatio-temporal supervision, given the predicted blind-spot maps and our generated pseudo maps, we employ a listwise loss to match their relative distributions, a cosine loss to align their overall response patterns, and a pairwise loss to rank high-utility locations above low-utility ones:
\begin{equation}
    \mathcal{L}_{\mathrm{map}}
    =
    \lambda_{\mathrm{list}}\mathcal{L}_{\mathrm{list}}
    +
    \lambda_{\mathrm{cos}}\mathcal{L}_{\mathrm{cos}}
    +
    \lambda_{\mathrm{pair}}\mathcal{L}_{\mathrm{pair}}.
    \label{eq:map_objective}
\end{equation}
At the clip level, we define $\mathcal{L}_{\mathrm{act}}$ as the binary cross-entropy loss between $q_i$ and the pseudo intervention label $y_i^{\mathrm{int}}$. We also introduce a preservation loss
$\mathcal{L}_{\mathrm{keep}}$ that minimizes $q_i$ for clips whose
visual predictions are already correct. Finally, we adopt the camera motion recognition loss $\mathcal{L}_{\mathrm{cls}}$ following previous methods. The complete objective is
\begin{equation}
    \mathcal{L}
    =
    \lambda_{\mathrm{cls}}\mathcal{L}_{\mathrm{cls}}
    +
    \lambda_{\mathrm{act}}\mathcal{L}_{\mathrm{act}}
    +
    \lambda_{\mathrm{keep}}\mathcal{L}_{\mathrm{keep}}
    +
    \mathcal{L}_{\mathrm{map}}.
    \label{eq:training_objective}
\end{equation}

\section{Experiments}
\label{sec:experiments}
\subsection{Experimental Setup}
\label{sec:experimental_setup}

\begin{table*}[!t]
\centering
\begingroup
\footnotesize
\setlength{\tabcolsep}{3.0pt}
\renewcommand{\arraystretch}{1.12}
\begin{tabular*}{\textwidth}{
    @{\extracolsep{\fill}}l
    ccc
    cc
    c@{}
}
\toprule
& \multicolumn{3}{c}{\textbf{CameraMotionDataset}}
& \multicolumn{2}{c}{\textbf{CameraBench}}
& \multicolumn{1}{c}{\textbf{ShotBench}} \\
\cmidrule(lr){2-4}
\cmidrule(lr){5-6}
\cmidrule(l){7-7}
\textbf{Model}
& \textbf{Inst. Acc.}$\uparrow$
& \textbf{Macro-F1}$\uparrow$
& \textbf{Weighted-F1}$\uparrow$
& \textbf{mAP}$\uparrow$
& \textbf{AB-Acc.}$\uparrow$
& \textbf{CM-Acc.}$\uparrow$ \\
\midrule

Qwen3-VL-8B-Instruct~\citep{bai2025qwen3vl}
& \textemdash & \textemdash & \textemdash & 35.94 & 73.54 & 44.83 \\
InternVL2.5-8B~\citep{chen2024internvl25}
& \textemdash & \textemdash & \textemdash & 29.50 & 59.00 & 33.60 \\

\midrule
VGGT camera cue~\citep{feng2026geometry}
& 73.80 & 87.00 & 92.00 & \textemdash & \textemdash & \textemdash \\
Cam-Motion-72B~\citep{lin2025camera}
& \textemdash & \textemdash & \textemdash & 62.80 & 80.60 & 56.47\\
CamReasoner-7B~\citep{wu2026camreasoner}
& \textemdash & \textemdash & \textemdash & 34.30 & 78.40 & 39.70 \\
ShotVL-7B~\citep{liu2025shotbench}
& \textemdash & \textemdash & \textemdash & 42.24 & 69.15 & 62.90 \\

\midrule
VGGT-1.1B~\citep{wang2025vggt}
& 79.07 & 91.80 & 93.32 & 25.92 & 88.11 & 54.96 \\
V-JEPA~2.1, 317M~\cite{murlabadia2026vjepa21}
& \underline{80.05}
& \underline{93.09}
& \underline{93.70}
& \underline{70.74}
& \underline{91.89}
& \underline{82.11} \\
\textbf{Fovea-JEPA (1.5B)}
& \textbf{83.47}
& \textbf{93.47}
& \textbf{94.50}
& \textbf{71.47}
& \textbf{92.37}
& \textbf{87.28} \\
\bottomrule
\end{tabular*}
\caption{
Performance of Fovea-JEPA against state-of-the-art methods on three benchmarks.
}
\label{tab:main_results}
\endgroup
\end{table*}

\paragraph{Datasets and metrics.}
We evaluate Fovea-JEPA on three camera-motion benchmarks, including CameraMotionDataset~\citep{feng2026geometry}, CameraBench~\citep{lin2025camera}, and ShotBench~\citep{liu2025shotbench}. CameraMotionDataset contains 12,274 clips
annotated with 15 camera-motion primitives. We follow its official split of 9,819 training, 1,227 validation, and 1,228 test clips, and report exact-match instance accuracy (\emph{Inst.~Acc.}), Macro-F1, and Weighted-F1. On CameraBench, we report mAP following the benchmark protocol, and also employ the average binary accuracy (AB-Acc.), which is defined as the unweighted mean of ordinary binary accuracy across the 15 motion tasks following CamReasoner~\citep{wu2026camreasoner}. For ShotBench, we use ShotQA for task adaptation and evaluate on the official 464-example camera-movement subset, reporting camera-movement accuracy (\emph{CM-Acc.}).

\paragraph{Baselines.}
We compare against general-purpose VLMs and task-specific camera-motion models. We consider Qwen3-VL-8B-Instruct~\citep{bai2025qwen3vl} and InternVL2.5-8B~\citep{chen2024internvl25} as the general-purpose baselines. The task-specific baselines include a VGGT camera-cue classifier~\citep{feng2026geometry}, Cam-Motion-72B~\citep{lin2025camera}, CamReasoner-7B~\citep{wu2026camreasoner}, and ShotVL-7B~\citep{liu2025shotbench}. We directly adopt the officially reported results of all baseline methods for fair comparison.

\paragraph{Implementation details.}
All our experiments are conducted on a single RTX 4090 GPU. We uniformly sample eight frames from each clip. The two branches use frozen V-JEPA~2.1 ViT-L~\cite{murlabadia2026vjepa21} and VGGT~\cite{wang2025vggt} backbones with input resolutions of $384\times384$ and $518\times518$, respectively. Five-fold cross-validation is conducted, and we use out-of-fold predictions to obtain intervention labels. Our model is trained for 12 epochs with AdamW using a batch size of 8, a learning rate of $1\times10^{-4}$, and weight decay of $1\times10^{-4}$. The weights for classification loss \(\mathcal{L}_{\text{cls}}\), action loss \(\mathcal{L}_{\text{act}}\), preservation loss \(\mathcal{L}_{\text{keep}}\), listwise loss \(\mathcal{L}_{\text{list}}\), cosine loss \(\mathcal{L}_{\text{cos}}\), and pairwise loss \(\mathcal{L}_{\text{pair}}\) are set to \(0.35\), \(0.25\), \(0.15\), \(0.10\), \(0.05\), and \(0.02\), respectively. For the clip-level intervention score, we retain the top $\xi_1=3\%$ of responses per frame and the top $\xi_2=75\%$ of frame scores.

\subsection{Main Results}
\label{sec:main_results}

Table~\ref{tab:main_results} compares Fovea-JEPA with published results from general-purpose VLMs and task-specific camera-motion models. We additionally include the constituent VGGT and V-JEPA~2.1 branches as controlled baselines. Fovea-JEPA outperforms all baselines consistently on three datasets. On CameraMotionDataset, Fovea-JEPA achieves 83.47\% Inst.~Acc., 93.47\% Macro-F1, and 94.50\% Weighted-F1, while the strongest baseline, the VGGT camera-cue classifier
~\citep{feng2026geometry}, achieves 73.80\%, 87.00\%, and 92.00\%, respectively. On CameraBench, Fovea-JEPA achieves 71.47\% mAP and 92.37\% AB-Acc., respectively. Fovea-JEPA also outperforms the strongest external baseline, Cam-Motion-72B~\citep{lin2025camera}, by \(8.67\%\) in mAP and \(11.77\%\) in AB-Acc. On ShotBench, the cutting-edge ShotVL-7B achieves \(62.90\%\) CM-Acc., while Fovea-JEPA outperform it by \(24.38\%\). These significant gains validate that JEPAs augmented with geometric intervention are a considerable solution for camera motion recognition.

\begin{figure*}[!t]
    \centering
    \includegraphics[width=\textwidth]
    {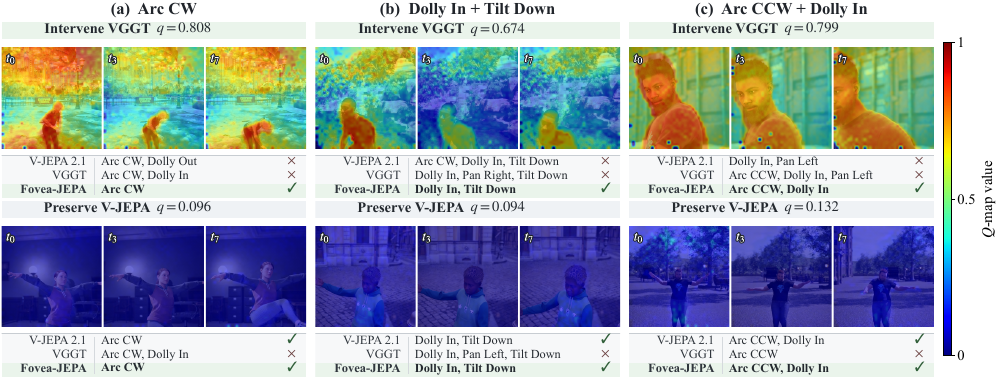}
    \caption{
    Qualitative analysis of the learned intervention decisions.
    Each column pairs two clips with the same ground-truth camera
    motion. For high-$q$ cases, $q>\tau$ activates the
    VGGT-guided intervention: the predictions from both V-JEPA~2.1
    and VGGT are inexact, whereas Fovea-JEPA recovers the exact
    label set. For low-$q$ cases, $q<\tau$ preserves the V-JEPA
    pathway, retaining its exact prediction despite an inexact
    VGGT prediction. }
    \label{fig:qmap_qualitative}
\end{figure*}

\subsection{Ablation Studies}
\label{sec:ablation}
We conduct extensive ablation studies on CameraMotionDataset to validate Fovea-JEPA thoroughly. More experimental results can be founded in the supplemental material.

\begin{table}[h]
\centering
{\small
\setlength{\tabcolsep}{3.5pt}
\begin{tabular}{@{}lrr@{}}
\toprule
Setting & V-JEPA~2.1 & VGGT \\
\midrule
w/o reversal
& 76.71 $\pm$ 0.53
& 71.44 $\pm$ 0.25 \\
w/ reversal
& 80.03 $\pm$ 0.61
& 78.72 $\pm$ 0.31 \\
\midrule
$\Delta$
& +3.32
& +7.28 \\
\bottomrule
\end{tabular}
}
\caption{Ablation of temporal-reversal supervision on
exact-match accuracy (\%). Results are reported as
mean $\pm$ sample standard deviation over three runs.}
\label{tab:temporal_reversal_ablation}
\end{table}
\subsubsection{Temporal Reversal and Geometric Blind Spots}
\label{sec:reversal_ablation}

We evaluate direction-aware temporal-reversal supervision,
which preserves each reversible motion family while
exchanging its directional labels. As shown in
Table~\ref{tab:temporal_reversal_ablation}, both branches
improve in overall exact-match accuracy, with VGGT exhibiting
the larger gain. The aggregate improvement, however, conceals different
family-level responses. For V-JEPA~2.1, pair accuracy
increases by \(4.00\) and \(1.16\) percentage points for roll
and dolly, but decreases by \(3.76\) and \(3.13\) points for
tilt and truck, respectively. In contrast, VGGT improves
across every evaluated motion family while remaining the
weaker standalone predictor. This asymmetric response
suggests that VGGT retains complementary geometric cues in
motion families for which V-JEPA~2.1 responds less
consistently. It therefore motivates retaining V-JEPA~2.1 as
the primary representation and using VGGT as a geometry
expert to compensate for geometric blind spots selectively.

\paragraph{Blind-Spot Localization.}
We first examine whether the token-level blind-spot map \(Q\) identifies important geometric regions. Under a matched \(50\%\) token budget, we retain VGGT locations ranked by \(Q\) before the fixed VGGT geometry head. Retaining the top-ranked half yields an exact-match accuracy of \(0.8143\), whereas retaining the bottom-ranked half reduces it to \(0.7622\). Random masks fall between the two ranked controls. This indicates that high-\(Q\) locations contain geometric evidence facilitating camera-motion prediction.

\begin{figure}[t]
    \centering
    \includegraphics[width=\columnwidth, height=5cm]
    {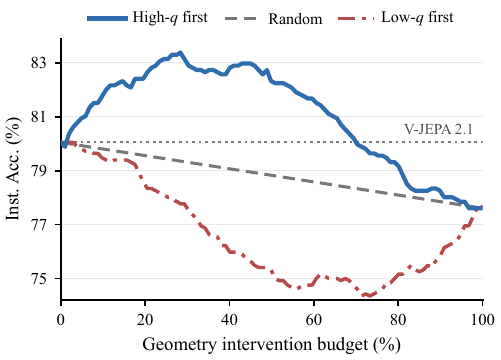}
    \caption{Clip-level intervention-budget analysis. The random curve is the expected
    same-budget accuracy.}
    \label{fig:intervention_budget}
\end{figure}

\paragraph{Clip-level routing.}
We next test whether aggregating \(Q\) into the clip-level
score \(q\) provides a reliable decision of when geometry
should intervene. The eligible test set contains 108 rescue
cases, for which geometric intervention corrects an
erroneous V-JEPA~2.1 prediction, and 983 clips already
classified correctly by V-JEPA~2.1. On this set, \(q\)
achieves an AUROC of \(0.855\) and an AUPRC of \(0.361\),
compared with a rescue prevalence of \(0.099\). Using the fixed intervention threshold $\tau$, intervening
on high-$q$ clips reaches \(0.8347\) exact-match accuracy
and yields a net correction of \(+42\) clips. Applying the same
intervention budget to low-\(q\) clips reduces accuracy to
\(0.7761\), while random routing reaches \(0.7937\).
Figure~\ref{fig:intervention_budget} extends this
comparison across intervention budgets. Descending-$q$
routing consistently outperforms ascending-$q$ and random
routing, indicating that $q$, rather than the budget alone,
determines which clips benefit from targeted geometric
intervention.

\paragraph{Qualitative analysis.}
Figure~\ref{fig:qmap_qualitative} contrasts high- and
low-\(q\) clips with the same ground-truth labels. In
high-\(q\) cases, blind-spot-guided intervention corrects
inexact predictions from both branches. Notably, although
neither V-JEPA~2.1 nor VGGT alone produces the exact label
set in these examples, Fovea-JEPA does. This result
illustrates the intended complementarity: VGGT need not be
the stronger standalone predictor to provide geometric cues
that are missing from V-JEPA~2.1 and enable a correct fused
prediction. In low-\(q\) cases,
it suppresses unnecessary intervention and preserves the
correct V-JEPA~2.1 prediction despite an inexact VGGT
output. The \(Q\)-maps show that the need for geometry
varies across locations and frames: \(Q\) localizes relevant
evidence, while \(q\) controls whether intervention is applied.

\paragraph{Objective ablation.}
Table~\ref{tab:main_loss_ablation} evaluates the main
training objectives. Without the action loss, no sample
crosses the intervention threshold and the model reduces to
the V-JEPA~2.1 baseline. Removing the classification loss
substantially decreases exact-match accuracy and net
correction, whereas removing the keep loss increases harmful
correct-to-wrong transitions. Removing all map-supervision
terms also degrades performance, demonstrating the importance
of spatial geometric supervision.

\begin{table}[h]
\centering
{\small
\setlength{\tabcolsep}{3.5pt}
\begin{tabular}{@{}lrrr@{}}
\toprule
Configuration & Inst. & C$\rightarrow$W & Net \\
\midrule
Full
& \textbf{83.47} & 59 & +42 \\
\textit{w/o} classification
& 81.68 & 54 & +20 \\
\textit{w/o} action
& 80.05 & 0 & 0 \\
\textit{w/o} keep
& 83.14 & 66 & +38 \\
\textit{w/o} map supervision
& 82.41 & 57 & +29 \\
\bottomrule
\end{tabular}
}
\caption{Ablation of the main training objectives.}
\label{tab:main_loss_ablation}
\end{table}

\section{Conclusion}
\label{sec:conclusion}

In this paper, we propose Fovea-JEPA, a dual-branch adaptive geometric intervention framework for video camera motion recognition. Our framework compensates for the geometric blind spots of JEPA models. Different from static fusion strategies that may introduce noisy geometric cues, our method leverages hierarchical pseudo-supervision with clip-level intervention labels and token-level geometric saliency targets to train a blind-spot predictor. This allows us to locate critical geometric regions and identify unreliable visual predictions, and eventually perform adaptive token-level feature modulation and clip-level threshold-based prediction correction. Extensive experiments validate our design, which effectively combines robust visual generalization and precise geometric structural priors.

\bibliography{aaai2027}

@String{CVPR    = {Proceedings of the IEEE/CVF Conference on Computer Vision and Pattern Recognition}}

@String{CVPRW   = {Proceedings of the IEEE/CVF Conference on Computer Vision and Pattern Recognition Workshops}}

@String{ECCV    = {Proceedings of the European Conference on Computer Vision}}

@String{ICCV    = {Proceedings of the IEEE/CVF International Conference on Computer Vision}}

@String{ICCVW   = {Proceedings of the IEEE/CVF International Conference on Computer Vision Workshops}}

@String{ICLR    = {International Conference on Learning Representations}}

@String{ICML    = {Proceedings of the International Conference on Machine Learning}}

@String{NEURIPS = {Advances in Neural Information Processing Systems}}

@String{TMLR    = {Transactions on Machine Learning Research}}

@inproceedings{feng2026geometry,
  author    = {Feng, Haoan and Musunuri, Sri Harsha and
               Su, Guan-Ming},
  title     = {Geometry-Guided Camera Motion Understanding in
               {VideoLLMs}},
  booktitle = CVPRW,
  pages     = {8423--8433},
  month     = {June},
  year      = {2026}
}

@inproceedings{lin2025camera,
  author    = {Lin, Zhiqiu and Cen, Siyuan and Jiang, Daniel and
               Karhade, Jay and Wang, Hewei and Mitra, Chancharik and
               Ling, Yu Tong Tiffany and Huang, Yuhan and
               Zawar, Rushikesh and Bai, Xue and Du, Yilun and
               Gan, Chuang and Ramanan, Deva},
  title     = {Towards Understanding Camera Motions in Any Video},
  booktitle = NEURIPS,
  volume    = {38},
  year      = {2025}
}

@misc{alrasheed2026latent,
  author        = {Alrasheed, Ali J. and Parast, Aryan Yazdan and
                   Azam, Basim and Bailey, James and Akhtar, Naveed},
  title         = {Latent Video Prediction Learns Better World Models},
  year          = {2026},
  eprint        = {2605.15618},
  archivePrefix = {arXiv}
}

@misc{murlabadia2026vjepa21,
  author        = {Mur-Labadia, Lorenzo and Muckley, Matthew and
                   Bar, Amir and Assran, Mahmoud and Sinha, Koustuv and
                   Rabbat, Michael and LeCun, Yann and Ballas, Nicolas and
                   Bardes, Adrien},
  title         = {{V-JEPA 2.1}: Unlocking Dense Features in Video
                   Self-Supervised Learning},
  year          = {2026},
  eprint        = {2603.14482},
  archivePrefix = {arXiv}
}

@inproceedings{wang2025vggt,
  author    = {Wang, Jianyuan and Chen, Minghao and Karaev, Nikita and
               Vedaldi, Andrea and Rupprecht, Christian and
               Novotny, David},
  title     = {{VGGT}: Visual Geometry Grounded Transformer},
  booktitle = CVPR,
  pages     = {5294--5306},
  month     = {June},
  year      = {2025}
}

@inproceedings{qian2021cvrl,
  author    = {Qian, Rui and Meng, Tianjian and Gong, Boqing and
               Yang, Ming-Hsuan and Wang, Huisheng and
               Belongie, Serge and Cui, Yin},
  title     = {Spatiotemporal Contrastive Video Representation
               Learning},
  booktitle = CVPR,
  pages     = {6964--6974},
  year      = {2021}
}

@inproceedings{jenni2021timeequivariant,
  author    = {Jenni, Simon and Jin, Hailin},
  title     = {Time-Equivariant Contrastive Video Representation
               Learning},
  booktitle = ICCV,
  pages     = {9970--9980},
  year      = {2021}
}

@inproceedings{feichtenhofer2022maest,
  author    = {Feichtenhofer, Christoph and Fan, Haoqi and
               Li, Yanghao and He, Kaiming},
  title     = {Masked Autoencoders as Spatiotemporal Learners},
  booktitle = NEURIPS,
  volume    = {35},
  pages     = {35946--35958},
  year      = {2022}
}

@inproceedings{tong2022videomae,
  author    = {Tong, Zhan and Song, Yibing and Wang, Jue and
               Wang, Limin},
  title     = {{VideoMAE}: Masked Autoencoders Are Data-Efficient
               Learners for Self-Supervised Video Pre-Training},
  booktitle = NEURIPS,
  volume    = {35},
  pages     = {10078--10093},
  year      = {2022}
}

@inproceedings{wang2023mvd,
  author    = {Wang, Rui and Chen, Dongdong and Wu, Zuxuan and
               Chen, Yinpeng and Dai, Xiyang and Liu, Mengchen and
               Yuan, Lu and Jiang, Yu-Gang},
  title     = {Masked Video Distillation: Rethinking Masked Feature
               Modeling for Self-Supervised Video Representation
               Learning},
  booktitle = CVPR,
  pages     = {6312--6322},
  year      = {2023}
}

@article{bardes2024vjepa,
  author  = {Bardes, Adrien and Garrido, Quentin and Ponce, Jean and
             Chen, Xinlei and Rabbat, Michael and LeCun, Yann and
             Assran, Mahmoud and Ballas, Nicolas},
  title   = {Revisiting Feature Prediction for Learning Visual
             Representations from Video},
  journal = TMLR,
  year    = {2024}
}

@misc{wang2025cinetech,
  author        = {Wang, Xinran and Xu, Songyu and Shan, Xiangxuan and
                   Zhang, Yuxuan and Diao, Muxi and Duan, Xueyan and
                   Huang, Yanhua and Liang, Kongming and Ma, Zhanyu},
  title         = {{CineTechBench}: A Benchmark for Cinematographic
                   Technique Understanding and Generation},
  year          = {2025},
  eprint        = {2505.15145},
  archivePrefix = {arXiv}
}

@inproceedings{liu2025shotbench,
  author    = {Liu, Hongbo and He, Jingwen and Jin, Yi and
               Zheng, Dian and Dong, Yuhao and Zhang, Fan and
               Huang, Ziqi and He, Yinan and Li, Yangguang and
               Chen, Weichao and Qiao, Yu and Ouyang, Wanli and
               Zhao, Shengjie and Liu, Ziwei},
  title     = {{ShotBench}: Expert-Level Cinematic Understanding
               in Vision-Language Models},
  booktitle = NEURIPS,
  volume    = {38},
  year      = {2025}
}

@misc{wu2026camreasoner,
  author        = {Wu, Hang and Cai, Yujun and Li, Zehao and Ge, Haonan and
                   Sun, Bowen and Yuan, Junsong and Wang, Yiwei},
  title         = {{CamReasoner}: Reinforcing Camera Movement Understanding
                   via Structured Spatial Reasoning},
  year          = {2026},
  eprint        = {2602.00181},
  archivePrefix = {arXiv}
}

@inproceedings{schonberger2016sfm,
  author    = {Sch{\"o}nberger, Johannes L. and
               Frahm, Jan-Michael},
  title     = {Structure-from-Motion Revisited},
  booktitle = CVPR,
  pages     = {4104--4113},
  year      = {2016}
}

@inproceedings{wang2024dust3r,
  author    = {Wang, Shuzhe and Leroy, Vincent and Cabon, Yohann and
               Chidlovskii, Boris and Revaud, Jerome},
  title     = {{DUSt3R}: Geometric {3D} Vision Made Easy},
  booktitle = CVPR,
  pages     = {20697--20709},
  year      = {2024}
}

@inproceedings{wang2024vggsfm,
  author    = {Wang, Jianyuan and Karaev, Nikita and
               Rupprecht, Christian and Novotny, David},
  title     = {{VGGSfM}: Visual Geometry Grounded Deep Structure
               From Motion},
  booktitle = CVPR,
  pages     = {21686--21697},
  year      = {2024}
}

@inproceedings{zhang2025monst3r,
  author    = {Zhang, Junyi and Herrmann, Charles and Hur, Junhwa and
               Jampani, Varun and Darrell, Trevor and Cole, Forrester
               and Sun, Deqing and Yang, Ming-Hsuan},
  title     = {{MonST3R}: A Simple Approach for Estimating Geometry
               in the Presence of Motion},
  booktitle = ICLR,
  year      = {2025}
}

@inproceedings{pickup2014arrow,
  author    = {Pickup, Lyndsey C. and Pan, Zheng and
               Wei, Donglai and Shih, YiChang and
               Zhang, Changshui and Zisserman, Andrew and
               Sch{\"o}lkopf, Bernhard and Freeman, William T.},
  title     = {Seeing the Arrow of Time},
  booktitle = CVPR,
  pages     = {2035--2042},
  year      = {2014}
}

@inproceedings{wei2018arrow,
  author    = {Wei, Donglai and Lim, Joseph J. and
               Zisserman, Andrew and Freeman, William T.},
  title     = {Learning and Using the Arrow of Time},
  booktitle = CVPR,
  pages     = {8052--8060},
  year      = {2018}
}

@inproceedings{price2019retro,
  author    = {Price, Will and Damen, Dima},
  title     = {Retro-Actions: Learning {Close} by
               Time-Reversing {Open} Videos},
  booktitle = ICCVW,
  pages     = {1371--1380},
  year      = {2019}
}

@inproceedings{jenni2020temporal,
  author    = {Jenni, Simon and Meishvili, Givi and
               Favaro, Paolo},
  title     = {Video Representation Learning by Recognizing
               Temporal Transformations},
  booktitle = ECCV,
  pages     = {425--442},
  year      = {2020}
}

@inproceedings{xue2025arrow,
  author    = {Xue, Zihui and Luo, Mi and Grauman, Kristen},
  title     = {Seeing the Arrow of Time in Large Multimodal
               Models},
  booktitle = NEURIPS,
  volume    = {38},
  year      = {2025}
}

@misc{han2026tracing,
  author        = {Han, Peitao and Cheng, Fei and
                   Pereira, Lis K. and Liu, Qianying and
                   Kitazawa, Shigeru},
  title         = {Tracing the Arrow of Time: Diagnosing
                   Temporal Information Flow in {Video-LLMs}},
  year          = {2026},
  eprint        = {2605.07568},
  archivePrefix = {arXiv},
  primaryClass  = {cs.CV}
}

@inproceedings{li2023blip2,
  author    = {Li, Junnan and Li, Dongxu and Savarese, Silvio and
               Hoi, Steven},
  title     = {{BLIP}-2: Bootstrapping Language-Image Pre-training
               with Frozen Image Encoders and Large Language Models},
  booktitle = ICML,
  series    = {Proceedings of Machine Learning Research},
  volume    = {202},
  pages     = {19730--19742},
  year      = {2023}
}

@inproceedings{assran2023ijepa,
  author    = {Assran, Mahmoud and Duval, Quentin and Misra, Ishan and
               Bojanowski, Piotr and Vincent, Pascal and Rabbat, Michael and
               LeCun, Yann and Ballas, Nicolas},
  title     = {Self-Supervised Learning From Images With a
               Joint-Embedding Predictive Architecture},
  booktitle = CVPR,
  pages     = {15619--15629},
  year      = {2023}
}

@misc{assran2025vjepa2,
  author        = {Assran, Mido and Bardes, Adrien and Fan, David and
                   Garrido, Quentin and Howes, Russell and Komeili, Mojtaba and
                   Muckley, Matthew and Rizvi, Ammar and Roberts, Claire and
                   Sinha, Koustuv and Zholus, Artem and others},
  title         = {{V-JEPA 2}: Self-Supervised Video Models Enable
                   Understanding, Prediction and Planning},
  year          = {2025},
  eprint        = {2506.09985},
  archivePrefix = {arXiv}
}

@misc{bai2025qwen3vl,
  title         = {{Qwen3-VL} Technical Report},
  author        = {Shuai Bai and Yuxuan Cai and Ruizhe Chen and Keqin Chen and
                   Xionghui Chen and Zesen Cheng and Lianghao Deng and Wei Ding and
                   Chang Gao and Chunjiang Ge and Wenbin Ge and Zhifang Guo and
                   Qidong Huang and Jie Huang and Fei Huang and Binyuan Hui and
                   Shutong Jiang and Zhaohai Li and Mingsheng Li and Mei Li and
                   Kaixin Li and Zicheng Lin and Junyang Lin and Xuejing Liu and
                   Jiawei Liu and Chenglong Liu and Yang Liu and Dayiheng Liu and
                   Shixuan Liu and Dunjie Lu and Ruilin Luo and Chenxu Lv and
                   Rui Men and Lingchen Meng and Xuancheng Ren and Xingzhang Ren and
                   Sibo Song and Yuchong Sun and Jun Tang and Jianhong Tu and
                   Jianqiang Wan and Peng Wang and Pengfei Wang and Qiuyue Wang and
                   Yuxuan Wang and Tianbao Xie and Yiheng Xu and Haiyang Xu and
                   Jin Xu and Zhibo Yang and Mingkun Yang and Jianxin Yang and
                   An Yang and Bowen Yu and Fei Zhang and Hang Zhang and Xi Zhang and
                   Bo Zheng and Humen Zhong and Jingren Zhou and Fan Zhou and
                   Jing Zhou and Yuanzhi Zhu and Ke Zhu},
  year          = {2025},
  eprint        = {2511.21631},
  archivePrefix = {arXiv},
  primaryClass  = {cs.CV},
  url           = {https://arxiv.org/abs/2511.21631}
}

@misc{chen2024internvl25,
  title         = {Expanding Performance Boundaries of Open-Source Multimodal
                   Models with Model, Data, and Test-Time Scaling},
  author        = {Zhe Chen and Weiyun Wang and Yue Cao and Yangzhou Liu and
                   Zhangwei Gao and Erfei Cui and Jinguo Zhu and Shenglong Ye and
                   Hao Tian and Zhaoyang Liu and Lixin Gu and Xuehui Wang and
                   Qingyun Li and Yiming Ren and Zixuan Chen and Jiapeng Luo and
                   Jiahao Wang and Tan Jiang and Bo Wang and Conghui He and
                   Botian Shi and Xingcheng Zhang and Han Lv and Yi Wang and
                   Wenqi Shao and Pei Chu and Zhongying Tu and Tong He and
                   Zhiyong Wu and Huipeng Deng and Jiaye Ge and Kai Chen and
                   Kaipeng Zhang and Limin Wang and Min Dou and Lewei Lu and
                   Xizhou Zhu and Tong Lu and Dahua Lin and Yu Qiao and
                   Jifeng Dai and Wenhai Wang},
  year          = {2025},
  eprint        = {2412.05271},
  archivePrefix = {arXiv},
  primaryClass  = {cs.CV},
  url           = {https://arxiv.org/abs/2412.05271}
}

\end{document}


\maketitle

This supplementary material complements the main paper with
exact definitions of the optimization objectives, additional
experimental results, and further details of the datasets and
evaluation protocols that could not be included because of
space constraints.

\appendix

\section{Detailed Optimization Objectives}
\label{sec:supp_objectives}

The main paper summarizes the optimization objective of
Fovea-JEPA and provides the weights of its constituent loss
terms. Here, we give their exact definitions and specify the
samples that contribute to each objective. Following the
notation of the main paper, let $B$ denote the mini-batch size
and $K$ the number of camera-motion classes. For the $i$-th
sample, $\mathbf y^i\in\{0,1\}^{K}$ denotes the ground-truth
multi-label vector, $q_i\in[0,1]$ denotes the clip-level
intervention score, and
$\widetilde{\mathbf z}^{i}\in\mathbb{R}^{K}$ denotes the
camera-motion logits produced after blind-spot-guided
geometric intervention.

\subsection{Clip-Level Objectives}
\label{sec:supp_clip_objectives}

\paragraph{Camera-motion classification loss.}
We supervise the intervened logits using multi-label binary
cross-entropy over all samples and motion classes:
\begin{equation}
\begin{aligned}
\mathcal{L}_{\mathrm{cls}}
=
-\frac{1}{BK}
\sum_{i=1}^{B}
\sum_{k=1}^{K}
\Big[
&y_k^i
\log \sigma\bigl(\widetilde z_k^i\bigr)
\\
&+
\left(1-y_k^i\right)
\log\left(
1-\sigma\bigl(\widetilde z_k^i\bigr)
\right)
\Big].
\end{aligned}
\label{eq:supp_cls_loss}
\end{equation}
All training samples contribute to
$\mathcal{L}_{\mathrm{cls}}$, including those that do not
receive action supervision.

\paragraph{Action loss.}
Following the intervention-label definition in the main
paper, $y_i^{\mathrm{int}}=1$ when the visual-branch
prediction $\widehat{\mathbf y}_j^i$ is incorrect and the
fused prediction $\widehat{\mathbf y}_f^i$ is correct, while
$y_i^{\mathrm{int}}=0$ when
$\widehat{\mathbf y}_j^i$ is correct. Samples for which both
predictions are incorrect receive no intervention
supervision. Let $\mathcal{I}_{\mathrm{act}}$ denote the set
of samples assigned an intervention label.

To balance the intervention and non-intervention cases, we
assign the sample weight
\begin{equation}
\omega_i
=
\begin{cases}
8, & y_i^{\mathrm{int}}=1,\\
1, & y_i^{\mathrm{int}}=0,
\end{cases}
\qquad
i\in\mathcal{I}_{\mathrm{act}}.
\label{eq:supp_action_weight}
\end{equation}
The weighted action loss is
\begin{equation}
\begin{aligned}
\mathcal{L}_{\mathrm{act}}
=
-\frac{1}
{\sum_{i\in\mathcal{I}_{\mathrm{act}}}\omega_i}
\sum_{i\in\mathcal{I}_{\mathrm{act}}}
\omega_i
\Big[
&y_i^{\mathrm{int}}\log q_i
\\
&+
\left(1-y_i^{\mathrm{int}}\right)
\log(1-q_i)
\Big].
\end{aligned}
\label{eq:supp_action_loss}
\end{equation}
When $\mathcal{I}_{\mathrm{act}}$ is empty,
$\mathcal{L}_{\mathrm{act}}$ is defined as zero.

\paragraph{Prediction-preservation loss.}
Let
\begin{equation}
\mathcal{I}_{\mathrm{keep}}
=
\left\{
i\in\mathcal{I}_{\mathrm{act}}
\;\middle|\;
y_i^{\mathrm{int}}=0
\right\}
\label{eq:supp_keep_set}
\end{equation}
denote the samples for which the primary V-JEPA prediction is
already exactly correct. Although these samples serve as
negative examples in $\mathcal{L}_{\mathrm{act}}$, we
additionally penalize their intervention scores:
\begin{equation}
\mathcal{L}_{\mathrm{keep}}
=
\frac{1}
{\left|\mathcal{I}_{\mathrm{keep}}\right|}
\sum_{i\in\mathcal{I}_{\mathrm{keep}}}
q_i.
\label{eq:supp_keep_loss}
\end{equation}
This objective suppresses unnecessary geometric intervention
on reliable V-JEPA predictions. If a mini-batch contains no
sample from $\mathcal{I}_{\mathrm{keep}}$, the preservation
loss is set to zero.

\subsection{Map-Level Objectives}
\label{sec:supp_map_objectives}

Let $\mathbf Q\in[0,1]^{T\times N}$ denote the predicted
blind-spot map and $u_{i,t,n}$ the token-level
geometric-saliency target. The map losses align their
relative spatiotemporal structures without constraining
absolute response magnitudes.

For notational convenience, let
$p\in\{1,\ldots,TN\}$ index the flattened
spatial-temporal positions of a sample. We denote the
flattened geometric-saliency target by $u_{i,p}$ and define
the corresponding pre-sigmoid blind-spot logit as
\begin{equation}
b_{i,p}
=
\operatorname{logit}(Q_{i,p})
=
\log\frac{Q_{i,p}}{1-Q_{i,p}}.
\label{eq:supp_map_logit}
\end{equation}
Before computing the map-level losses, the predicted logits
and saliency targets are standardized independently within
each sample:
\begin{equation}
b'_{i,p}
=
\frac{b_{i,p}-\mu_i^{b}}
{\max(s_i^{b},\epsilon)},
\qquad
u'_{i,p}
=
\frac{u_{i,p}-\mu_i^{u}}
{\max(s_i^{u},\epsilon)},
\label{eq:supp_map_standardization}
\end{equation}
where $\mu_i^{b}$ and $\mu_i^{u}$ are the corresponding
within-sample means, $s_i^{b}$ and $s_i^{u}$ are the
within-sample standard deviations, and
$\epsilon=10^{-5}$.

To give greater weight to samples with more informative
saliency variation, we define
\begin{equation}
\rho_i
=
\operatorname{clip}
\left(
\frac{s_i^{u}}
{\frac{1}{B}\sum_{j=1}^{B}s_j^{u}},
\,0.1,\,
5.0
\right).
\label{eq:supp_map_sample_weight}
\end{equation}
The weights $\rho_i$ are shared across all three map-level
objectives.

\paragraph{Listwise distribution loss.}
We convert the standardized saliency targets and predicted
map logits into spatial-temporal distributions:
\begin{equation}
\begin{aligned}
\pi_{i,p}^{u}
&=
\frac{
\exp\left(u'_{i,p}/\tau_u\right)
}{
\sum_{r=1}^{TN}
\exp\left(u'_{i,r}/\tau_u\right)
},
\\[3pt]
\pi_{i,p}^{Q}
&=
\frac{
\exp\left(b'_{i,p}/\tau_Q\right)
}{
\sum_{r=1}^{TN}
\exp\left(b'_{i,r}/\tau_Q\right)
}.
\end{aligned}
\label{eq:supp_listwise_distributions}
\end{equation}
where $\tau_u=\tau_Q=0.5$. The listwise objective minimizes
the Kullback--Leibler divergence from the saliency
distribution to the predicted map distribution:
\begin{equation}
\mathcal{L}_{\mathrm{list}}
=
\frac{
\sum_{i=1}^{B}
\rho_i
D_{\mathrm{KL}}
\left(
\boldsymbol{\pi}_i^{u}
\middle\|
\boldsymbol{\pi}_i^{Q}
\right)
}{
\sum_{i=1}^{B}\rho_i
}.
\label{eq:supp_listwise_loss}
\end{equation}
This objective encourages $\mathbf Q$ to allocate high
relative probability to spatial-temporal positions with high
geometric saliency.

\paragraph{Cosine alignment loss.}
To align the overall response pattern without constraining
its magnitude, we minimize the cosine distance between the
standardized map logits and saliency targets:
\begin{equation}
\mathcal{L}_{\mathrm{cos}}
=
\frac{
\sum_{i=1}^{B}
\rho_i
\left[
1-
\frac{
{\mathbf b'_i}^{\top}\mathbf u'_i
}{
\lVert\mathbf b'_i\rVert_2
\lVert\mathbf u'_i\rVert_2
}
\right]
}{
\sum_{i=1}^{B}\rho_i
}.
\label{eq:supp_cosine_loss}
\end{equation}

\paragraph{Pairwise ranking loss.}
For each sample, we identify the positions with the highest
and lowest standardized saliency:
\begin{equation}
p_i^{+}
=
\arg\max_{p} u'_{i,p},
\qquad
p_i^{-}
=
\arg\min_{p} u'_{i,p}.
\label{eq:supp_saliency_extrema}
\end{equation}
The pairwise objective requires the predicted response at
$p_i^{+}$ to exceed that at $p_i^{-}$ by a margin $m$:
\begin{equation}
\mathcal{L}_{\mathrm{pair}}
=
\frac{
\sum_{i=1}^{B}
\rho_i
\max
\left(
0,\,
m-b'_{i,p_i^{+}}+b'_{i,p_i^{-}}
\right)
}{
\sum_{i=1}^{B}\rho_i
},
\label{eq:supp_pairwise_loss}
\end{equation}
where $m=0.05$.

The complete map-level objective is
\begin{equation}
\mathcal{L}_{\mathrm{map}}
=
\lambda_{\mathrm{list}}\mathcal{L}_{\mathrm{list}}
+
\lambda_{\mathrm{cos}}\mathcal{L}_{\mathrm{cos}}
+
\lambda_{\mathrm{pair}}\mathcal{L}_{\mathrm{pair}},
\label{eq:supp_complete_map_loss}
\end{equation}

\section{Additional Experimental Results}
\label{sec:supp_additional_results}

This section presents additional experimental results for
Fovea-JEPA. We first analyze temporal-reversal supervision
across motion families and then report a complete ablation
of the optimization objectives.

\subsection{Direction-Aware Temporal Reversal}
\label{sec:supp_reversal}

The main paper reports the aggregate effects of
temporal-reversal supervision on the two representation
pathways. We further examine how these effects vary across motion
families.

\subsubsection{Family-Level Effects of Temporal Reversal}
\label{sec:supp_family_effects}

The family-level results reveal a marked asymmetry between
the two representation pathways under temporal-reversal
supervision. Table~\ref{tab:supp_reversal_family} compares
the corresponding accuracy changes across motion families
for V-JEPA~2.1 and VGGT.

\begin{table}[t]
\centering
{\small
\setlength{\tabcolsep}{7.5pt}
\begin{tabular}{@{}lcc@{}}
\toprule
& \multicolumn{2}{c}{$\Delta$ Acc. (pp)} \\
\cmidrule(lr){2-3}
Motion family & V-JEPA~2.1 & VGGT \\
\midrule
Roll       & $+4.00$ & $+5.78$ \\
Dolly      & $+1.16$ & $+3.57$ \\
Arc        & $\phantom{+}0.00$ & $+0.71$ \\
Pedestal   & $-0.54$ & $+2.45$ \\
Truck      & $-3.13$ & $+3.61$ \\
Tilt       & $-3.76$ & $+1.34$ \\
Pan        & $-6.99$ & $+1.40$ \\
Static     & -- & -- \\
\bottomrule
\end{tabular}
}
\caption{Family-level accuracy changes under
temporal-reversal supervision. Values are percentage-point
differences relative to training without reversal. Static is omitted because its label is invariant under
temporal reversal.}
\label{tab:supp_reversal_family}
\end{table}

For V-JEPA~2.1, temporal reversal improves accuracy for roll
and dolly, leaves arc unchanged, and causes negative transfer
for pedestal, truck, tilt, and pan. VGGT, in contrast, shows
positive mean changes for all seven reversible motion
families, including the four families for which V-JEPA~2.1
deteriorates.

This asymmetry is particularly informative because VGGT
remains the weaker standalone predictor. Its positive
response indicates that useful camera-motion-related
geometric cues remain available in motion families for which
V-JEPA~2.1 does not exploit them reliably. Consistent with
the main paper, we interpret this representation asymmetry as
evidence of geometric blind spots: spatial regions whose
camera-motion-relevant geometry is underrepresented by
V-JEPA~2.1 but remains accessible to the geometry pathway.
This observation motivates retaining V-JEPA~2.1 as the
primary representation while using VGGT to provide selective
geometric compensation.

\subsection{Complete Objective Ablation}
\label{sec:supp_objective_ablation}

The main paper reports the principal objective ablations.
Table~\ref{tab:supp_complete_objective_ablation} provides
the complete component-wise evaluation. Each variant removes
only the indicated objective from the full training loss;
removing map supervision eliminates
$\mathcal{L}_{\mathrm{list}}$, $\mathcal{L}_{\cos}$, and
$\mathcal{L}_{\mathrm{pair}}$ jointly.

\begin{table}[t]
\centering
{\small
\setlength{\tabcolsep}{3.2pt}
\begin{tabular}{@{}lrrrrr@{}}
\toprule
Configuration
& Inst.
& Macro
& Wtd.
& C$\rightarrow$W
& Net \\
\midrule
Full
& 83.47 & 93.47 & 94.50 & 59 & $+42$ \\
\textit{w/o} classification
& 81.68 & 93.85 & 94.50 & 54 & $+20$ \\
\textit{w/o} action
& 80.05 & 93.09 & 93.70 & 0 & $\phantom{+}0$ \\
\textit{w/o} keep
& 83.14 & 93.35 & 94.41 & 66 & $+38$ \\
\textit{w/o} listwise
& 82.33 & 93.42 & 94.26 & 65 & $+28$ \\
\textit{w/o} cosine
& 83.14 & 93.39 & 94.41 & 60 & $+38$ \\
\textit{w/o} pairwise
& 83.14 & 93.33 & 94.32 & 62 & $+38$ \\
\textit{w/o} map supervision
& 82.41 & 93.44 & 94.20 & 57 & $+29$ \\
\bottomrule
\end{tabular}
}
\caption{Ablation of the main training objectives. Inst.,
Macro, and Wtd. denote Exact Match, Macro-F1, and
Weighted-F1, respectively; all three are reported as
percentages. C$\rightarrow$W counts predictions changed
from exactly correct under V-JEPA~2.1 to incorrect after
intervention, while Net is W$\rightarrow$C minus
C$\rightarrow$W.}
\label{tab:supp_complete_objective_ablation}
\end{table}

The full objective achieves the highest Exact Match of
$83.47\%$ and the largest net correction of $+42$. Removing
$\mathcal{L}_{\mathrm{act}}$ prevents all clips from crossing
the intervention threshold, reducing the model to the
V-JEPA~2.1 baseline. Removing $\mathcal{L}_{\mathrm{cls}}$
decreases Exact Match by $1.79$ percentage points and reduces
the net correction from $+42$ to $+20$. Although its
Macro-F1 is slightly higher and its Weighted-F1 remains
unchanged, the lower Exact Match and Net values show that
$\mathcal{L}_{\mathrm{cls}}$ is important for recovering the
complete label set. Removing $\mathcal{L}_{\mathrm{keep}}$
increases C$\rightarrow$W transitions from $59$ to $66$,
confirming its role in suppressing unnecessary intervention
on reliable V-JEPA~2.1 predictions.

Among the individual map-level objectives,
$\mathcal{L}_{\mathrm{list}}$ has the largest effect:
removing it decreases Exact Match to $82.33\%$ and Net to
$+28$. Removing either $\mathcal{L}_{\cos}$ or
$\mathcal{L}_{\mathrm{pair}}$ produces a smaller but
consistent reduction in Exact Match and net correction.
Removing all map supervision results in an Exact Match of
$82.41\%$ and a Net value of $+29$. Although the number of
C$\rightarrow$W transitions decreases slightly from $59$ to
$57$, the substantially lower Net value indicates fewer
successful W$\rightarrow$C corrections. The three map-level
objectives therefore provide complementary supervision for
learning blind-spot structures that support effective
geometric intervention. Their individual effects need not be
additive because they jointly optimize the same predicted
map $\mathbf{Q}$.

\section{Datasets and Evaluation Protocols}
\label{app:datasets}

\subsection{CameraMotionDataset Protocol}
\label{sec:supp_cameramotion_protocol}

CameraMotionDataset~\citep{feng2026geometry} contains
12,274 non-overlapping one-second video clips generated
from synthetic scenes with known camera trajectories. We
use its official split of 9,819 training, 1,227 validation,
and 1,228 test clips.

Each clip is annotated with a multi-hot vector over 15
camera-motion primitives. The label space contains seven
directional pairs: arc (clockwise/counterclockwise), crane
(up/down), dolly (in/out), pan (left/right), roll
(clockwise/counterclockwise), tilt (up/down), and truck
(left/right), together with a static class. In the
family-level analyses, we refer to the dataset's crane pair
as the pedestal family.

Let $\mathcal{D}_{\mathrm{test}}$ denote the test set, and let
$\mathbf{y}_i,\widehat{\mathbf{y}}_i\in\{0,1\}^{K}$ denote
the ground-truth and predicted label vectors for clip $i$,
where $K=15$. We report Instance Accuracy (Exact Match),
Macro-F1, and Weighted-F1:
\begin{align}
\mathrm{Inst.\ Acc.}
&=
\frac{1}{|\mathcal{D}_{\mathrm{test}}|}
\sum_{i\in\mathcal{D}_{\mathrm{test}}}
\mathbf{1}
\bigl[
\widehat{\mathbf{y}}_i=\mathbf{y}_i
\bigr],
\\
\mathrm{Macro\mbox{-}F1}
&=
\frac{1}{K}
\sum_{c=1}^{K}
\mathrm{F1}_c,
\\
\mathrm{Weighted\mbox{-}F1}
&=
\frac{\sum_{c=1}^{K}n_c\mathrm{F1}_c}
{\sum_{c=1}^{K}n_c},
\end{align}
where
$n_c=\sum_{i\in\mathcal{D}_{\mathrm{test}}}y_{i,c}$ is the
number of positive test instances for class $c$. All three
metrics are reported as percentages. Instance Accuracy
requires the complete predicted label set to match the
ground truth exactly.

Following the constraint-aware decoding protocol of
CameraMotionDataset~\citep{feng2026geometry}, class
probabilities are thresholded at $0.5$ and processed in
descending score order. Opposite directions within the same
motion family are treated as mutually exclusive, and the
static class cannot co-occur with a dynamic primitive. We
retain at most three predicted labels per clip. If no class
exceeds the threshold, the highest-scoring class is
selected. The same decoding procedure is applied to VGGT,
V-JEPA~2.1, and Fovea-JEPA.

\subsection{CameraBench Protocol}
\label{sec:supp_camerabench_protocol}

CameraBench~\citep{lin2025camera} contains approximately
3,000 real-world internet videos with expert annotations of
geometric and semantic camera motions. We evaluate the 15
camera-centric motion tasks: dolly in/out, pedestal up/down,
truck left/right, zoom in/out, pan left/right, tilt up/down,
roll clockwise/counterclockwise, and static.

CameraBench evaluates camera-motion understanding across
15 motion tasks. Following the official evaluation protocol,
we evaluate 7,679 valid video--motion decisions from the
test set and report task-wise average precision and binary
accuracy.

Because CameraBench videos can be longer than the clips in
CameraMotionDataset, we use at most eight temporal windows
per video. If more windows are available, eight are randomly
sampled during training and selected at evenly spaced
positions during validation and testing. Let
\(\mathcal{W}_v\) denote the selected windows for video
\(v\). Window-level class logits are aggregated by
element-wise maximum:
\begin{equation}
z_{v,c}
=
\max_{w\in\mathcal{W}_v} z_{w,c}.
\end{equation}
For Fovea-JEPA, the video-level intervention score is
similarly obtained as
\begin{equation}
q_v
=
\max_{w\in\mathcal{W}_v} q_w.
\end{equation}
The same logit aggregation is used for all internally
controlled methods.

Following the original CameraBench protocol, average
precision is computed independently for each motion task
from the continuous video-level confidence scores. Let
\(K=15\) denote the number of camera-motion tasks. We report
their unweighted mean:
\begin{equation}
\mathrm{mAP}
=
\frac{1}{K}
\sum_{c=1}^{K}
\mathrm{AP}_c.
\end{equation}
Because average precision is computed from continuous
scores, it does not require a classification threshold.

To compare with the binary-classification results reported
by CamReasoner~\cite{wu2026camreasoner}, we use the same
category-wise aggregation and report Average Binary
Accuracy. A prediction is positive when its sigmoid
probability is at least \(0.5\). Let
\(\mathcal{V}_c\) denote the set of videos with a valid
annotation for motion task \(c\). The metric is the
unweighted mean of the \(K\) per-task binary accuracies:
\begin{equation}
\mathrm{Avg.\ Binary\ Acc.}
=
\frac{1}{K}\sum_{c=1}^{K}
\frac{1}{|\mathcal{V}_c|}
\sum_{v\in\mathcal{V}_c}
\mathbf{1}\bigl[
\widehat{y}_{v,c}=y_{v,c}
\bigr].
\label{eq:supp_camerabench_binary_accuracy}
\end{equation}

\subsection{ShotQA and ShotBench Protocol}
\label{sec:supp_shotbench_protocol}

ShotBench~\cite{liu2025shotbench} evaluates cinematic
understanding across eight dimensions. Its Camera Movement
subset contains 464 video-based multiple-choice questions.
Because Fovea-JEPA targets camera-motion recognition, we
report results on this subset rather than the aggregate
score over all ShotBench dimensions.

We train the task-specific components on the
camera-movement video portion of ShotQA. After retaining
examples with valid videos and camera-movement answers, the
training and validation splits contain 784 and 139 videos,
respectively.

We map the correct answer text of each ShotQA training
question to an 18-class canonical motion vocabulary:
push in, pull out, move left/right, pan left/right, tilt
up/down, boom up/down, zoom in/out, dolly zoom, arc,
tracking, camera roll, rack focus, and static. Equivalent
expressions are mapped to the same class; for example,
\emph{trucking left} and \emph{move left} share one
canonical label. Answers that describe multiple camera
motions produce multi-hot targets.

As with CameraBench, we use at most eight temporal windows
per video. Additional windows are randomly sampled during
training and selected at evenly spaced positions during
validation and evaluation. Window-level class logits and
intervention scores are aggregated by element-wise maximum
over the selected windows, following the aggregation
described in Sec.~\ref{sec:supp_camerabench_protocol}.

For ShotBench evaluation, each answer option is mapped to
one or more classes in the canonical motion vocabulary and
scored by the highest predicted probability among those
classes. We select the option with the highest score and
report CM Accuracy as the percentage of correctly answered
questions among the 464 Camera Movement examples. The same
decoding procedure is used for VGGT, V-JEPA~2.1, and
Fovea-JEPA.

\subsection{Statistical Testing Protocol}
\label{sec:supp_statistical_testing}

All statistical comparisons pair Fovea-JEPA and
V-JEPA~2.1 predictions on the same evaluation examples.
For CameraMotionDataset, we apply a two-sided exact McNemar
test to Exact Match correctness, using the numbers of
wrong-to-correct and correct-to-wrong predictions as the two
discordant counts.

We additionally construct a \(95\%\) confidence interval for
the Exact Match improvement using 10,000 paired bootstrap
trials. In each trial, the 1,228 test clips are sampled with
replacement, and the Exact Match difference between
Fovea-JEPA and V-JEPA~2.1 is recomputed on the same
resampled clips. The interval is given by the \(2.5\)th and
\(97.5\)th percentiles of the resulting distribution.

For CameraBench, bootstrap resampling is performed at the
video level, with all valid motion annotations of a sampled
video retained together. Average Binary Accuracy is
recomputed for both methods on every paired resample. For
ShotBench, the 464 Camera Movement questions are resampled
as paired evaluation items, and the difference in CM
Accuracy is recomputed. For both benchmarks, the reported confidence interval is
the corresponding \(95\%\) percentile interval. These paired analyses characterize
evaluation-sample uncertainty at the reported operating
points; variation across the three CameraMotionDataset
training runs is reported separately using the sample
standard deviation.

\subsection{Data Access and Usage Terms}
\label{sec:supp_data_access}

CameraMotionDataset~\cite{feng2026geometry} is obtained
from its official public release and used according to the
accompanying license and preprocessing instructions.
CameraBench~\citep{lin2025camera} training material is
access controlled and was obtained through the dataset
authors' official application procedure. The ShotQA training data are obtained through the official
gated-access procedure associated with
ShotBench~\cite{liu2025shotbench}. We follow the provider's
usage terms and use the official Camera Movement subset for
ShotBench evaluation.

\begin{table*}[!t]
\begin{minipage}{\textwidth}
\centering
{\small
\setlength{\tabcolsep}{3pt}
\renewcommand{\arraystretch}{1.12}
\begin{tabularx}{\textwidth}{@{}l*{9}{Y}@{}}
\toprule
Method
& \shortstack{Dolly\\In/Out}
& \shortstack{Pedestal\\Up/Down}
& \shortstack{Truck\\Right/Left}
& \shortstack{Zoom\\In/Out}
& \shortstack{Pan\\Right/Left}
& \shortstack{Tilt\\Up/Down}
& \shortstack{Roll\\CW/CCW}
& Static
& Avg. \\
\midrule

\multicolumn{10}{l}{
    \emph{SfM/SLAM and 3D reconstruction}
} \\
COLMAP
& 36.2/13.1
& 11.9/19.7
& 34.1/30.0
& 13.9/14.2
& 43.9/46.4
& 28.3/19.1
& 42.1/48.7
& 7.5
& 27.3 \\
VGGSfM
& 56.6/28.9
& 28.7/38.2
& 48.9/35.3
& 21.7/17.3
& 60.9/58.7
& 46.6/43.3
& 61.4/55.5
& 16.7
& 41.3 \\
MASt3R
& 47.5/21.1
& 23.5/40.2
& 37.8/38.1
& 42.2/46.6
& 66.6/58.0
& 63.2/40.3
& 50.4/53.5
& 15.7
& 43.1 \\
MegaSAM
& 73.8/43.2
& 24.2/29.1
& 45.3/44.2
& 11.1/10.2
& 79.5/82.2
& 73.8/65.3
& 71.5/75.8
& 22.0
& 50.1 \\
\midrule

\multicolumn{10}{l}{
    \emph{Zero-shot multimodal models}
} \\
GPT-4o
& 66.3/29.2
& 21.1/38.2
& 38.0/21.9
& 41.7/39.3
& 44.7/42.1
& 43.6/35.5
& 24.0/28.7
& 32.0
& 36.4 \\
InternVL3-78B
& 72.0/18.2
& 19.6/32.5
& 33.8/29.4
& 26.4/33.4
& 47.2/53.5
& 47.8/40.3
& 27.6/25.0
& 22.6
& 36.8 \\
Qwen2.5-VL-7B
& 56.0/14.9
& 18.7/30.5
& 34.5/27.6
& 29.8/43.4
& 62.7/66.7
& 54.5/34.1
& 18.8/24.2
& 19.8
& 35.8 \\
Qwen2.5-VL-72B
& 58.0/16.8
& 20.6/32.5
& 36.4/29.5
& 31.7/45.4
& 64.7/68.6
& 56.4/36.0
& 20.6/26.1
& 21.6
& 37.7 \\
Qwen3-VL-8B-Instruct
& 67.6/14.4
& 16.1/26.5
& 28.3/32.5
& 29.6/39.6
& 47.2/44.7
& 55.6/54.8
& 22.0/27.3
& 33.0
& 35.9 \\
\midrule

\multicolumn{10}{l}{
    \emph{Task-specific models}
} \\
Cam-Motion-7B
& 83.9/38.6
& 27.8/47.8
& 67.9/50.0
& 54.5/75.8
& 79.2/83.8
& 76.3/67.6
& 32.3/41.0
& 73.6
& 60.0 \\
Cam-Motion-32B
& 85.6/40.1
& 29.3/49.4
& 69.6/51.5
& 56.0/77.3
& \underline{80.7}/\underline{85.4}
& \underline{77.9}/69.2
& 33.9/42.7
& \underline{75.4}
& 61.6 \\
Cam-Motion-72B
& 86.8/41.3
& 30.5/50.6
& 70.7/52.6
& 57.1/78.5
& \textbf{81.9}/\textbf{86.6}
& \textbf{79.1}/\underline{70.4}
& 35.0/43.8
& \textbf{76.6}
& 62.8 \\
ShotVL-7B
& 61.2/16.0
& 15.6/28.4
& 44.0/36.8
& 33.5/51.9
& 70.6/77.7
& 63.8/55.5
& 27.4/26.8
& 24.4
& 42.2 \\
\midrule

VGGT
& 64.1/15.3
& 25.1/26.5
& 29.3/27.9
& 14.1/15.9
& 24.4/31.8
& 28.4/22.0
& 23.1/25.1
& 15.8
& 25.9 \\
V-JEPA~2.1
& \underline{87.8}/\underline{63.9}
& \textbf{62.1}/\underline{78.7}
& \underline{73.5}/\textbf{83.0}
& \underline{66.5}/\underline{82.3}
& 40.0/53.4
& 73.1/\textbf{72.0}
& \underline{87.6}/\textbf{90.8}
& 46.3
& \underline{70.7} \\
Fovea-JEPA
& \textbf{89.0}/\textbf{70.7}
& \underline{56.1}/\textbf{80.1}
& \textbf{73.5}/\underline{82.5}
& \textbf{69.4}/\textbf{83.5}
& 41.0/53.8
& 73.5/70.1
& \textbf{88.5}/\underline{90.1}
& 50.2
& \textbf{71.5} \\
\bottomrule
\end{tabularx}
}
\captionof{table}{Per-motion average precision (\%) on CameraBench.
For paired motion families, the two values in each cell follow
the directional order shown in the column header. Best and
second-best results are shown in bold and underlined,
respectively.}
\label{tab:supp_camerabench_map_full}
\end{minipage}
\end{table*}

\begin{table*}[!t]
\begin{minipage}{\textwidth}
\centering
{\small
\setlength{\tabcolsep}{3pt}
\renewcommand{\arraystretch}{1.12}
\begin{tabularx}{\textwidth}{@{}l*{9}{Y}@{}}
\toprule
Method
& \shortstack{Dolly\\In/Out}
& \shortstack{Pedestal\\Up/Down}
& \shortstack{Truck\\Right/Left}
& \shortstack{Zoom\\In/Out}
& \shortstack{Pan\\Right/Left}
& \shortstack{Tilt\\Up/Down}
& \shortstack{Roll\\CW/CCW}
& Static
& Avg. \\
\midrule

\multicolumn{10}{l}{
    \emph{Proprietary model}
} \\
GPT-4o
& 63.6/61.6
& 67.2/53.5
& 69.6/59.4
& 76.0/57.1
& 64.5/76.2
& 63.1/77.0
& 60.0/74.5
& 74.7
& 66.5 \\
\midrule

\multicolumn{10}{l}{
    \emph{Open-source multimodal models}
} \\
LLaVA-Video-7B
& 35.5/46.1
& 31.4/53.5
& 48.9/41.2
& 48.4/22.6
& 15.2/21.9
& 20.3/23.7
& 44.7/59.6
& 32.9
& 37.7 \\
InternVL2.5-8B
& 44.3/55.2
& 61.9/66.0
& 46.4/66.0
& 61.8/65.3
& 63.9/55.3
& 68.0/64.5
& 63.6/50.9
& 52.0
& 59.0 \\
Qwen2.5-VL-3B
& 64.7/61.6
& 84.7/81.9
& 71.3/79.9
& 85.1/72.3
& 77.9/79.5
& 84.9/89.7
& 71.7/75.5
& 17.8
& 69.3 \\
InternVL2.5-26B
& 62.1/50.8
& 75.3/80.5
& 74.0/72.8
& 58.0/61.5
& 72.0/71.4
& 82.0/78.8
& 66.8/65.2
& 71.8
& 69.5 \\
Qwen2.5-VL-7B
& 62.7/56.8
& 82.8/82.8
& 75.8/79.7
& 58.7/68.6
& 72.9/80.1
& 83.8/86.0
& 67.9/73.8
& 74.2
& 73.8 \\
Qwen2.5-VL-32B
& 73.6/76.1
& 80.2/80.2
& 72.5/65.2
& 67.6/73.9
& 78.5/78.9
& 71.6/66.7
& 89.4/78.3
& 78.0
& 75.4 \\
Qwen3-VL-8B-Instruct
& 61.0/59.9
& 77.9/81.9
& 62.8/69.3
& 71.8/79.0
& 66.8/71.5
& 81.0/80.9
& 82.3/78.3
& 78.7
& 73.5 \\
InternVL3-8B-Instruct
& 66.8/76.7
& 88.8/79.8
& 69.8/80.3
& 52.4/69.0
& 69.3/71.3
& 83.0/80.7
& 58.1/83.6
& 25.6
& 70.4 \\
\midrule

\multicolumn{10}{l}{
    \emph{Task-specific models}
} \\
Cam-Motion-7B
& 39.7/48.6
& 32.6/64.2
& 50.5/54.5
& 51.3/14.2
& 17.6/18.0
& 12.0/20.8
& 47.2/75.1
& 35.0
& 38.8 \\
Cam-Motion-72B
& 69.4/\underline{79.2}
& 80.0/79.5
& 77.6/77.5
& 81.8/80.3
& 80.5/80.7
& 84.7/87.5
& 84.5/\underline{82.3}
& 82.8
& 80.6 \\
CamReasoner-7B
& 68.7/57.7
& 84.9/80.5
& 84.6/83.2
& 74.2/72.8
& \textbf{85.7}/86.1
& 87.6/85.7
& 75.5/70.2
& 80.1
& 78.4 \\
ShotVL-7B
& 64.3/53.0
& 77.7/80.0
& 60.8/61.7
& 81.1/86.9
& 60.0/61.1
& 81.7/77.6
& 64.0/66.6
& 60.8
& 69.2 \\
\midrule

VGGT
& 79.6/90.3
& 92.1/91.4
& 84.2/88.5
& 87.8/88.9
& 85.0/84.6
& 87.3/92.3
& 90.4/89.1
& \textbf{90.0}
& 88.1 \\
V-JEPA~2.1
& \underline{88.7}/\underline{92.5}
& \underline{94.2}/\textbf{95.8}
& \underline{89.7}/\underline{93.6}
& \underline{92.2}/\textbf{96.2}
& 83.8/\textbf{88.5}
& \textbf{93.2}/\textbf{95.8}
& \textbf{97.7}/\textbf{96.8}
& 79.4
& \underline{91.9} \\
Fovea-JEPA
& \textbf{88.9}/\textbf{93.4}
& \textbf{94.4}/\underline{95.6}
& \textbf{91.0}/\textbf{94.7}
& \textbf{92.7}/\textbf{96.2}
& \underline{85.4}/\underline{87.5}
& \textbf{93.2}/\underline{95.4}
& \underline{95.1}/\underline{96.4}
& \underline{85.7}
& \textbf{92.4} \\
\bottomrule
\end{tabularx}
}
\captionof{table}{Per-motion binary accuracy (\%) on CameraBench.
For paired motion families, the two values in each cell follow
the directional order shown in the column header. Best and
second-best results are shown in bold and underlined,
respectively.}
\label{tab:supp_camerabench_acc_full}
\end{minipage}
\end{table*}

\section{Detailed Benchmark Results}
\label{app:benchmark_details}

This section complements the aggregate comparisons in the
main paper with benchmark-specific breakdowns. We report
per-motion results on CameraBench and a broader comparison
of Camera Movement accuracy on ShotBench, revealing how the
overall gains are distributed across motion categories and
model families.

\subsection{CameraBench Per-Motion Results}
\label{app:camerabench_details}

Tables~\ref{tab:supp_camerabench_map_full} and
\ref{tab:supp_camerabench_acc_full} disaggregate the
CameraBench results across its 15 camera-motion tasks.
Published baseline rows are reproduced from
CameraBench~\citep{lin2025camera}; VGGT, V-JEPA~2.1, and
Fovea-JEPA are evaluated using the protocol described in
Sec.~\ref{sec:supp_camerabench_protocol}.

\paragraph{Per-motion average precision.}
Table~\ref{tab:supp_camerabench_map_full} shows that
Fovea-JEPA achieves the highest average precision, improving
the mean AP from $70.7\%$ for V-JEPA~2.1 to $71.5\%$.
The improvement occurs on 10 of the 15 tasks and is largest
for dolly out and static, with gains of $6.8$ and $3.9$
percentage points, respectively. VGGT reaches an average AP
of $25.9\%$, substantially below both V-JEPA~2.1 and
Fovea-JEPA, indicating that its geometric representation is
not a stronger standalone ranking model on this benchmark.

\paragraph{Per-motion binary accuracy.}
Table~\ref{tab:supp_camerabench_acc_full} reports the
corresponding binary accuracies. Fovea-JEPA improves the
unweighted task average from $91.9\%$ to $92.4\%$, with the
largest gain on static, from $79.4\%$ to $85.7\%$. VGGT
obtains $88.1\%$ average binary accuracy despite its much
lower AP. This difference reflects the class imbalance in
CameraBench, with binary accuracy and AP providing different
perspectives on performance.

\FloatBarrier

\subsection{ShotBench Camera-Movement Results}
\label{app:shotbench_details}

Table~\ref{tab:supp_shotbench_cm_full} compares Camera
Movement (CM) accuracy on ShotBench. Published VideoLLM and
ShotVL results are reproduced from
ShotBench~\citep{liu2025shotbench}, and the CamReasoner result
is reproduced from its original evaluation
~\citep{wu2026camreasoner}. Among the published baselines,
ShotVL-7B obtains the highest accuracy of $62.90\%$.
V-JEPA~2.1 reaches $82.11\%$, while Fovea-JEPA further
improves the result to $87.28\%$, corresponding to gains of
$5.17$ percentage points over V-JEPA~2.1 and $24.38$ points
over ShotVL-7B. VGGT obtains $54.96\%$ when used alone,
consistent with its role as a complementary geometry pathway
rather than the primary predictor.

\begin{table}[H]
\centering
{\small
\setlength{\tabcolsep}{8pt}
\begin{tabular}{@{}lr@{}}
\toprule
Method & CM Accuracy \\
\midrule
\multicolumn{2}{l}{\emph{General-purpose VLMs}} \\
Qwen2.5-VL-3B-Instruct
& 31.90 \\
Qwen2.5-VL-7B-Instruct
& 30.20 \\
InternVL2.5-8B
& 33.60 \\
Qwen3-VL-8B-Instruct
& 44.83 \\
InternVL3-8B-Instruct
& 44.20 \\
GPT-4o
& 48.30 \\
Qwen2.5-VL-72B-Instruct
& 48.90 \\
\midrule
\multicolumn{2}{l}{\emph{Task-specific models}} \\
CamReasoner-7B
& 39.70 \\
Cam-Motion-7B
& 46.77 \\
ShotVL-3B
& 51.70 \\
ShotVL-7B
& 62.90 \\
\midrule
VGGT
& 54.96 \\
V-JEPA~2.1
& \underline{82.11} \\
Fovea-JEPA
& \textbf{87.28} \\
\bottomrule
\end{tabular}
}
\caption{Camera-movement (CM) accuracy (\%) on ShotBench.
Best and second-best results are shown in bold and
underlined, respectively.}
\label{tab:supp_shotbench_cm_full}
\end{table}

\bibliography{aaai2027}